\documentclass[10pt,twocolumn,letterpaper]{article}

\usepackage{iccv}
\usepackage{times}
\usepackage{epsfig}
\usepackage{graphicx}
\usepackage{amsmath}
\usepackage{amssymb}

\usepackage{multirow}
\usepackage{subcaption}
\usepackage{booktabs}
\usepackage{caption}
\usepackage{subcaption}
\usepackage[accsupp]{axessibility}
\graphicspath{{./imgs/}} 

\usepackage[pagebackref=true,breaklinks=true,letterpaper=true,colorlinks,bookmarks=false]{hyperref}

\iccvfinalcopy 

\def\iccvPaperID{11} 

\ificcvfinal\fi
\newcommand*{\affaddr}[1]{#1} 
\newcommand*{\affmark}[1][*]{\textsuperscript{#1}}

\begin{document}

\title{MILA: Multi-Task Learning from Videos via Efficient Inter-Frame Attention}
\author{%
  Donghyun Kim\affmark[1], Tian Lan\affmark[2], Chuhang Zou\affmark[2], Ning Xu\affmark[2], 
  \\Bryan A. Plummer\affmark[1], Stan Sclaroff\affmark[1], Jayan Eledath\affmark[2], Gerard Medioni\affmark[2]  \\ 
  \affaddr{\affmark[1]Boston University}, \affaddr{\affmark[2] Amazon}\\
  \tt\small $^1$\{donhk, bplum, sclaroff\}@bu.edu, $^2$\{tianlan, ninxu, zouchuha, eledathj, medioni\}@amazon.com}

\maketitle
\ificcvfinal\thispagestyle{empty}\fi

\begin{abstract}
   Prior work in multi-task learning has mainly focused on predictions on a single image. In this work, we present a new approach for multi-task learning from videos via efficient inter-frame local attention (MILA). Our approach contains a novel inter-frame attention module which allows learning of task-specific attention across frames. We embed the attention module in a ``slow-fast'' architecture, where the slow network runs on sparsely sampled keyframes and the fast shallow network runs on non-keyframes at a high frame rate. We also propose an effective adversarial learning strategy to encourage the slow and fast network to learn similar features to well align keyframes and non-keyframes. Our approach ensures low-latency multi-task learning while maintaining high quality predictions. MILA obatins competitive accuracy compared to state-of-the-art on two multi-task learning benchmarks while reducing the number of floating point operations (FLOPs) by up to 70\%. In addition, our attention based feature propagation method   (ILA) outperforms prior work in terms of task accuracy while also reducing up to 90\% of FLOPs.
\end{abstract}

\section{Introduction}

\begin{figure}[t]
		\centering
		\includegraphics[width=0.90\linewidth]{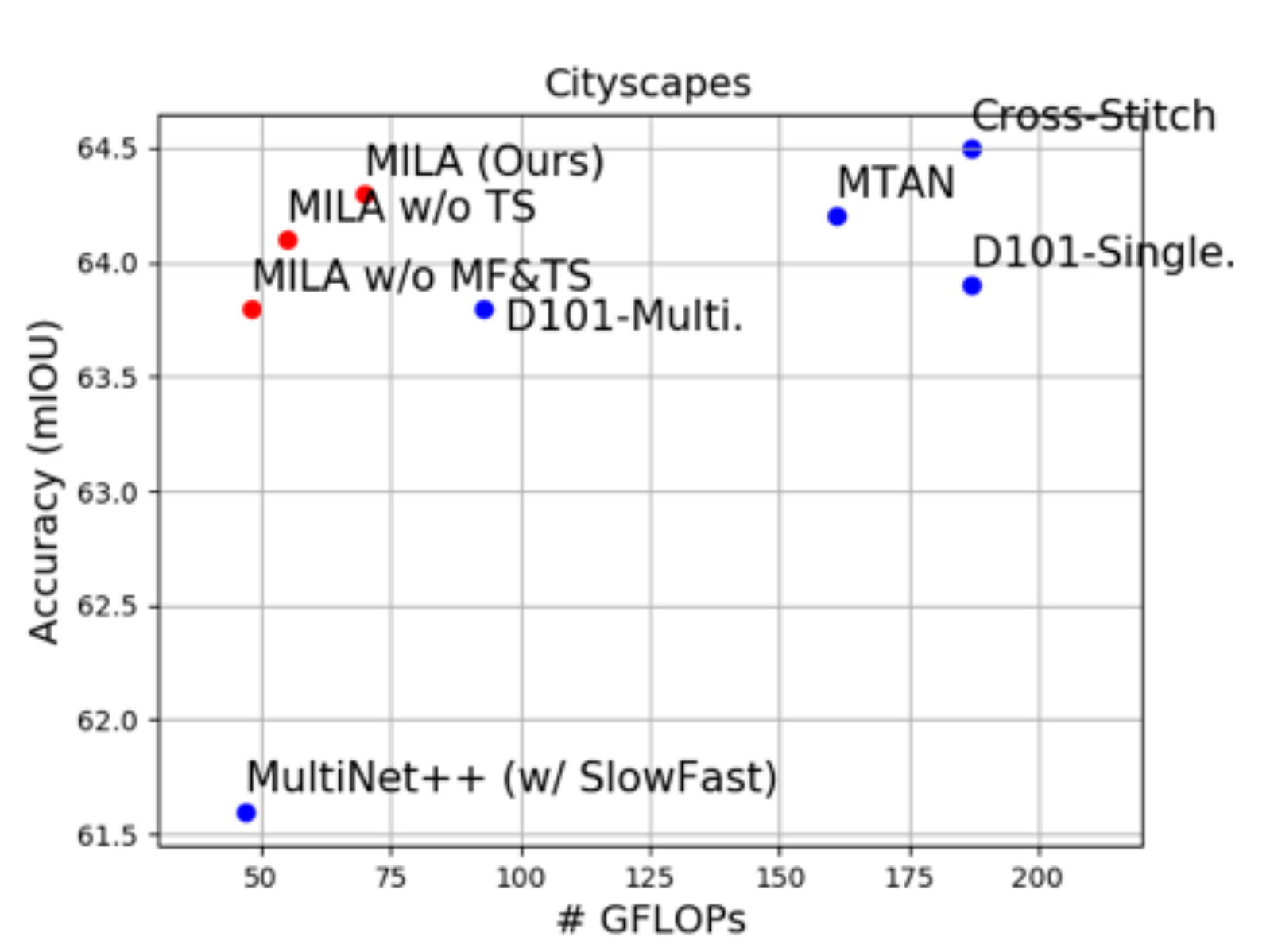} 
		\vskip -0.05in
		\caption{Comparison of the number of GFLOPs and mIOU performance of multi-task learning methods on Cityscapes. Our method (MILA) reduces computational burden significantly while maintaining accuracy. Our ILA module can be extended to attend task-specific (TS) and multi-frame features (MF) with minimal computations. We use the notation of each compared method and baseline from Sec.~\ref{exp:dataset}.}
		\label{fig:plot}
\end{figure}

Many computer vision applications, such as autonomous driving and indoor navigation, require multi-task predictions from video streams (\eg,~\cite{chen2018multi,chennupati2019multinet++,chowdhuri2019multinet}). For example, a self-driving system needs semantic segmentation at each time frame to understand what entities are around the car, and depth estimation to determine how far away each entity is.
This makes multi-task learning methods ideal since their shared representation can boost performance on each task while also being more computationally efficient.

\begin{figure*}[t]
		\centering
		\includegraphics[width=0.75\linewidth]{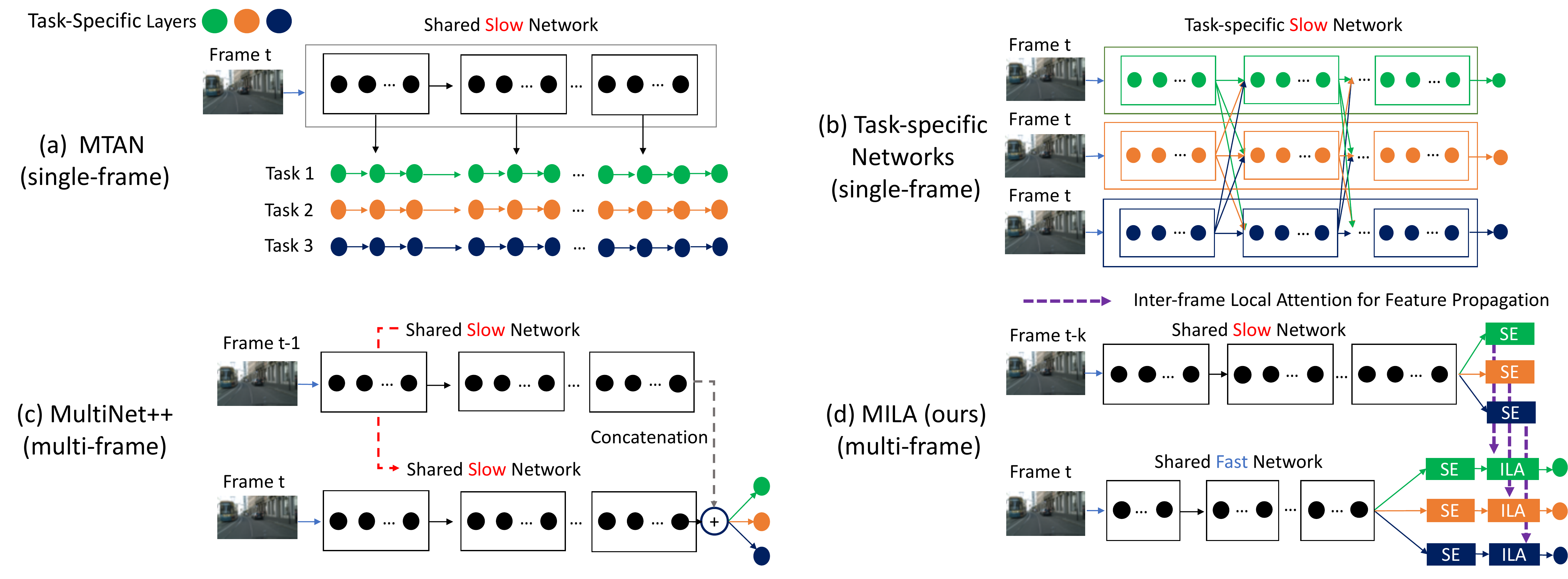} \\
		\caption{An illustration of difference between previous multi-task learning models and our multi-task model, MILA. Previous methods use only a $Slow$ network (\eg~ResNet101) for every frame ((a), (b), and (c)) and heavy task-specific layers for each task ((a) and (b)), which requires massive computations. In (d), we propose an efficient approach for multi-task learning from videos by utilizing a $Fast$ network (\eg~ResNet18) for non-keyframes and propagating the previous keyframe features from the $Slow$ network via our inter-frame local attention module (ILA). ILA is light-weight, accurate, and extended task-specific attention modules without requiring massive computations.
		}
		\label{fig:overview}
\end{figure*}

In this paper, we focus on efficient multi-task learning for dense pixel-wise predictions~(\eg~semantic segmentation and depth estimation) by leveraging a monocular video. Figure~\ref{fig:plot} compares the tradeoff between performance and computational burden for existing multi-task learning methods (blue) and our method (red).  Recent multi-task learning approaches for dense predictions primarily use single-frame predictions~\cite{gao2019nddr,kendall2018multi,liu2019end,misra2016cross} and often require heavy task-specific layers (illustrated in Figure~\ref{fig:overview}(a-b)), or a naive concatenation of the features from two consecutive video frames~\cite{chennupati2019multinet++} (Figure~\ref{fig:overview}(c)), which require massive amounts of floating point operations (FLOPs) to compute). To address this, we propose MILA -- multi-task learning framework for videos that exploits temporal cues using inter-frame local attention (ILA) modules as shown in Figure~\ref{fig:overview}(d).  Existing attention modules only attend to features in the current (single) frame~\cite{fu2019dual,liu2019end}, but ILA efficiently learns to attend and propagate features from previous frames. ILA is far more efficient than the expensive optical-flow based feature warping, which is widely used in previous work~\cite{jain2019accel,zhu2017deep}. In addition, the performance of optical flow warping based methods can be affected by the quality of estimated optical flow, which may fail on fast motion or occluded objects. 

Our MILA architecture embeds the ILA feature propagation method in a \textit{SlowFast} framework~\cite{feichtenhofer2019slowfast,jain2019accel}, which can reduce computational cost while maintaining comparable accuracy. In the \textit{SlowFast} architecture, keyframes are processed by a deep ($Slow$) network, and non-keyframes are processed by a shallow ($Fast$) network. 
Moreover, unlike the previous task-specific heavy layers, we show improvements in accuracy with our light-weight task-specific attention based ILA by leveraging temporal cues, and a novel adversarial learning strategy that encourages similar feature representations for both the $Slow$  and $Fast$  networks. Our ILA module differs from other attention-based approaches (\eg~\cite{fu2019dual}) in that we use attention to estimate the importance of features from different networks rather than the same network, which is a challenging problem that is not fully addressed by the existing attention modules.  Figure~\ref{fig:overview} illustrates the difference between our approach (MILA) and existing multi-task learning methods.

We evaluate our approach on two standard multi-task learning benchmarks: Cityscapes~\cite{cordts2016cityscapes} with outdoor scenes and NYUd v2~\cite{silberman2012indoor} with indoor scenes. As shown in Figure~\ref{fig:plot}, MILA method achieves on-par accuracy compared to the state-of-the-art multi-task learning methods, while reducing the number of FLOPs by up to 70\%.  MILA reduces the computational burden by a large margin without compromising accuracy. Moreover, we show that the ILA module can be used as a standalone feature propagation method in videos: it is much faster compared to existing feature propagation methods, and more accurate than the state-of-the-art~\cite{jain2019accel,li2018low} on semantic video segmentation.

Our contributions are: 
\begin{itemize}
    \item We address the task of video-based multi-task learning, which is not well explored in previous work. We present a multi-task learning framework via inter-frame local attention (MILA) that achieves competitive accuracy as compared to the state-of-the-art with a largely reduced computational cost.
    \item We introduce a new inter-frame local attention module (ILA) which learns task-specific features across frames. Our network is trained end-to-end with an adversarial loss for the \textit{SlowFast} network.
    \item Our ILA module can be used as a standalone feature propagation method in video tasks such as semantic segmentation, achieving the top accuracy with up to 90\% reduction of FLOPs.
\end{itemize}
     

\section{Related Work}
\noindent\textbf{Multi-task learning (MTL)} has shown improved accuracy or increased memory-efficiency for various tasks such as object classification, object detection and segmentation~\cite{chen2019hybrid,chen2018masklab,he2017mask,misra2016cross,strezoski2019many}, joint scene geometry and semantic segmentation~\cite{chennupati2019multinet++,eigen2015predicting,kendall2018multi,kokkinos2017ubernet,liu2019end,teichmann2018multinet,xu2018pad,zhou2020pattern}. Several methods are proposed to learn useful task-specific representations from shared representations or representations from other tasks~\cite{bragman2019stochastic,misra2016cross,zhou2020pattern,zhang2019pattern}. The cross-stitch unit~\cite{misra2016cross} links representations between different tasks. Zhou~\etal~\cite{zhou2020pattern} propose a pattern-structure diffusion for propagating inter and intra task-specific representations. However, previous approaches on dense prediction tasks (\eg, semantic segmentation) mainly focus on predictions from a single image.  Chennupati~\etal~\cite{chennupati2019multinet++} learn from videos by concatenating the features from two consecutive frames. Contrary to prior work, we go beyond single-frame based prediction and learn from videos by aggregating and propagating features across multiple frames.

 Although the shared representation of MTL can help improve generalization and reduce computational costs, it is also shown to potentially hurt accuracy due to the trade-off learning from multiple tasks~\cite{maninis2019attentive,standley2020tasks}. Kendall~\etal~\cite{kendall2018multi} use homeostatic uncertainty to weight different tasks adaptively during training. Other methods introduce complex task-specific layers~(\eg task-specific backbone)~\cite{liu2019end,misra2016cross,ruder122019latent} that also significantly increases the computational burden. In contrast, we show that our lightweight task-specific model design for our inter-frame attention module is able to achieve competitive task accuracy at a much lower computational cost via video learning. 
 \smallskip

\label{related:featurepropgation}
\noindent\textbf{Feature propagation} has been widely used in video applications to exploit temporal cues across frames~\cite{chennupati2019multinet++,gadde2017semantic,jain2019accel,li2018low,nilsson2018semantic,zhu2017deep} in order to reduce computational costs. Jain~\etal~\cite{jain2019accel} reduce the inference cost by combining the predictions of two network branches: a deep reference branch that computes detailed features from keyframes, and a shallower update branch that incorporates less detailed features at each frame with the wrapped features from a recently met keyframe. This is similar to the \textit{SlowFast}~\cite{feichtenhofer2019slowfast} design for video recognition. However, optical flow based feature propagation~\cite{jain2019accel} increases the computational cost with limited improvements in accuracy compared to the simple concatenation of features~\cite{chennupati2019multinet++}. \cite{jiang2019video,li2018low} use spatially variant convolution layers (SVC)  for feature propagation which is faster than optical flow warping. Our network, MILA, stems from the spirit of the \textit{SlowFast} network, and we use our light-weight inter-frame local attention (ILA) module for feature propagation instead of the expensive optical flow based approach. We also perform dense feature propagation between every neighboring frame, in addition to the sparse propagation between keyframes and non-keyframes only.  
\smallskip

\noindent\textbf{Attention modules} are widely used in various tasks such as natural language processing~(NLP)~\cite{deng2018latent,vaswani2017attention}, semantic segmentation~\cite{fu2019dual,huang2019ccnet,li2019attention,vaswani2017attention,zhang2018self}, image classification~\cite{hu2018squeeze,woo2018cbam,wang2020eca}, and video object detection~\cite{guo2019progressive,wang2018non,xiao2018video}. Channel-wise attention in CNNs has been proposed in~\cite{hu2018squeeze,wang2020eca}. Vaswani~\etal~\cite{vaswani2017attention} propose a self-attention module for a translation task by extracting global dependencies from input sequences. The self-attention first computes feature representations for query, key, and value, then computes global attention weights by measuring the similarity between the query and key. The final value can be obtained by a weighted sum of values from the sequence of input. Some video object detection methods use global/local attention modules~\cite{guo2019progressive,wang2018non,xiao2018video} for inter-frame features. These global/local spatial attention modules will not work well in our framework as we need to attend between two different representations (from $Slow$ and $Fast$ networks). In our work, we also add an adversarial loss in our training scheme (Sec~\ref{sec:mimicking}) to facilitate the attention module for the \textit{SlowFast} network and improve performance.

 \begin{figure}[t]
		\centering
		\includegraphics[width=0.85\linewidth]{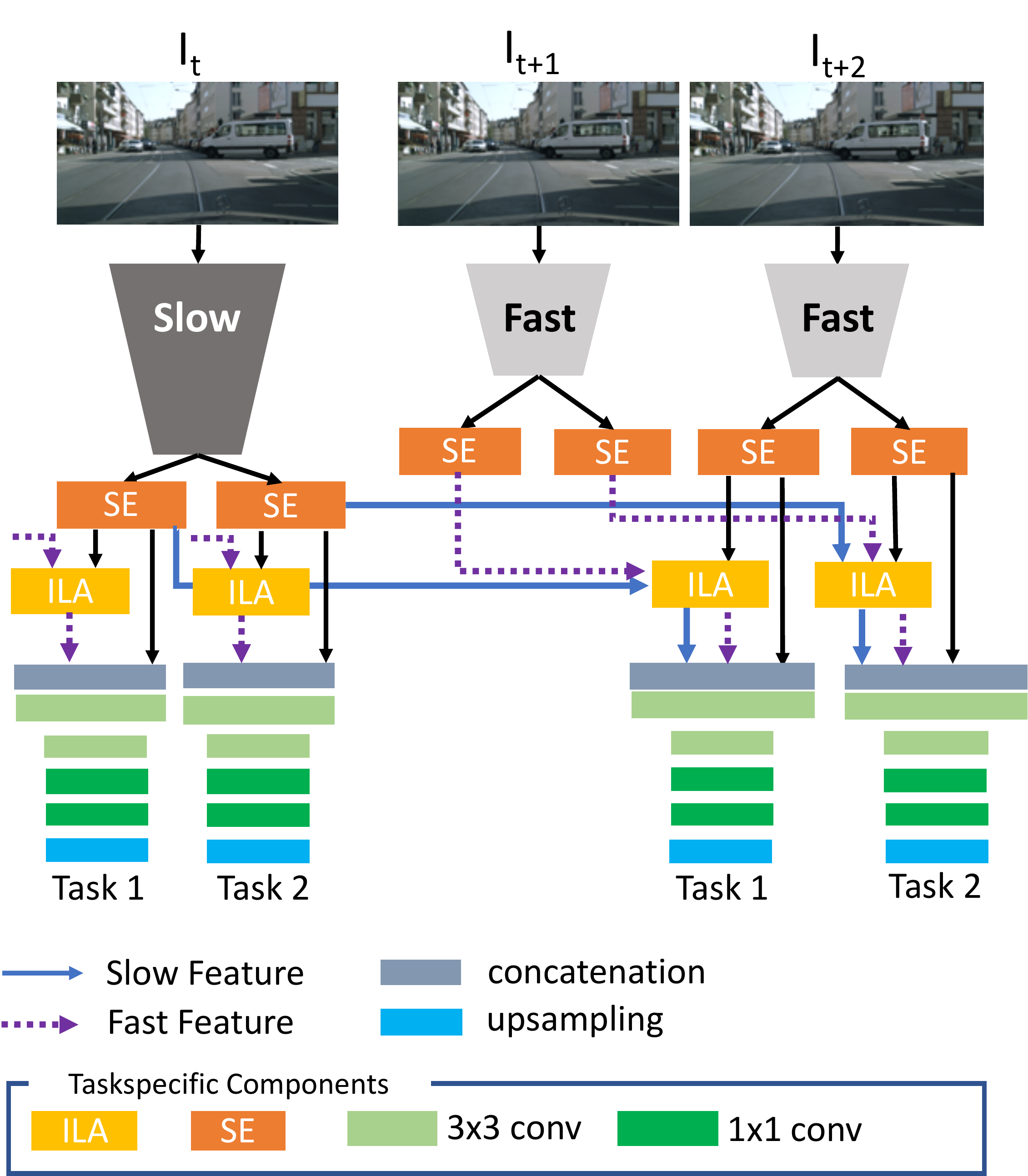} \\
		\caption{An illustration of our network architecture and the inference step of a keyframe $I_{t}$ and a non-keyframe $I_{t+2}$. ILA propagates multi-frame features to the current frame.}		\label{fig:architecture}
	    \vskip -0.2in
\end{figure}


\section{Multi-task Learning via Inter-Frame Local Attention}
We propose Multi-task Learning via Inter-Frame Local Attention (MILA), which is a computationally efficient multi-task learning model, with high quality prediction by leveraging temporal cues in video streams. Effectively learning spatial and temporal cues of different tasks in a light-weight and efficient manner is a challenge in real-time video applications.  Inspired by the \textit{SlowFast} network~\cite{feichtenhofer2019slowfast,jain2019accel}, we build an efficient multi-task network with a two-branch design: the $Slow$ branch runs on sparsely sampled keyframes and the light-weight $Fast$ network runs on non-keyframes. Unlike previous work that relies on heavy task-specific layers~\cite{liu2019end,misra2016cross}, we introduce a new light-weight task-specific attention module to learn and propagate task-specific features across frames. Figure~\ref{fig:architecture} shows the architecture of the proposed network.


In the following, we first explain our multi-task network architecture (MILA) in Sec.~\ref{sec:architecture}. Then, we introduce our novel task-specific inter-frame local attention (ILA) module in Sec.~\ref{sec:attention}, and an adversarial loss that further boosts the overall performance in Sec.~\ref{sec:mimicking}.

\subsection{Architecture Overview}
\label{sec:architecture}
MILA consists of two components: 1) a shared encoder network: a $Slow$ network that operates on sparsely sampled keyframes; a $Fast$ network runs on other frames. 2) $M$ task-specific decoder networks, one for each task. Each decoder network learns to attend to task-specific features from previous frames and the current frame (in Figure~\ref{fig:architecture}). 

The input is a sequence of $N$ RGB frames $I = \{I_1, I_2, \ldots, I_N\}$ from a monocular video, and the output is pixel-level predictions on $M$ tasks, $Y = \{y_1, y_2, ..., y_M\}$. At each time step $t\in\{1,2,\ldots,T\}$, we encode frame $I_t$ using the $Slow$  network if it is a keyframe, and the $Fast$ network otherwise. In our implementation, we use ResNet-101 as the $Slow$  network and ResNet-18 as the $Fast$  network. The encoder is shared among all tasks. We use $Slow(I_t)$ to denote features encoded by the $Slow$  network and $Fast(I_t)$ for features encoded by the $Fast$ network.

\begin{figure}[t]
    
        
    \centering
    \includegraphics[width=0.85\linewidth]{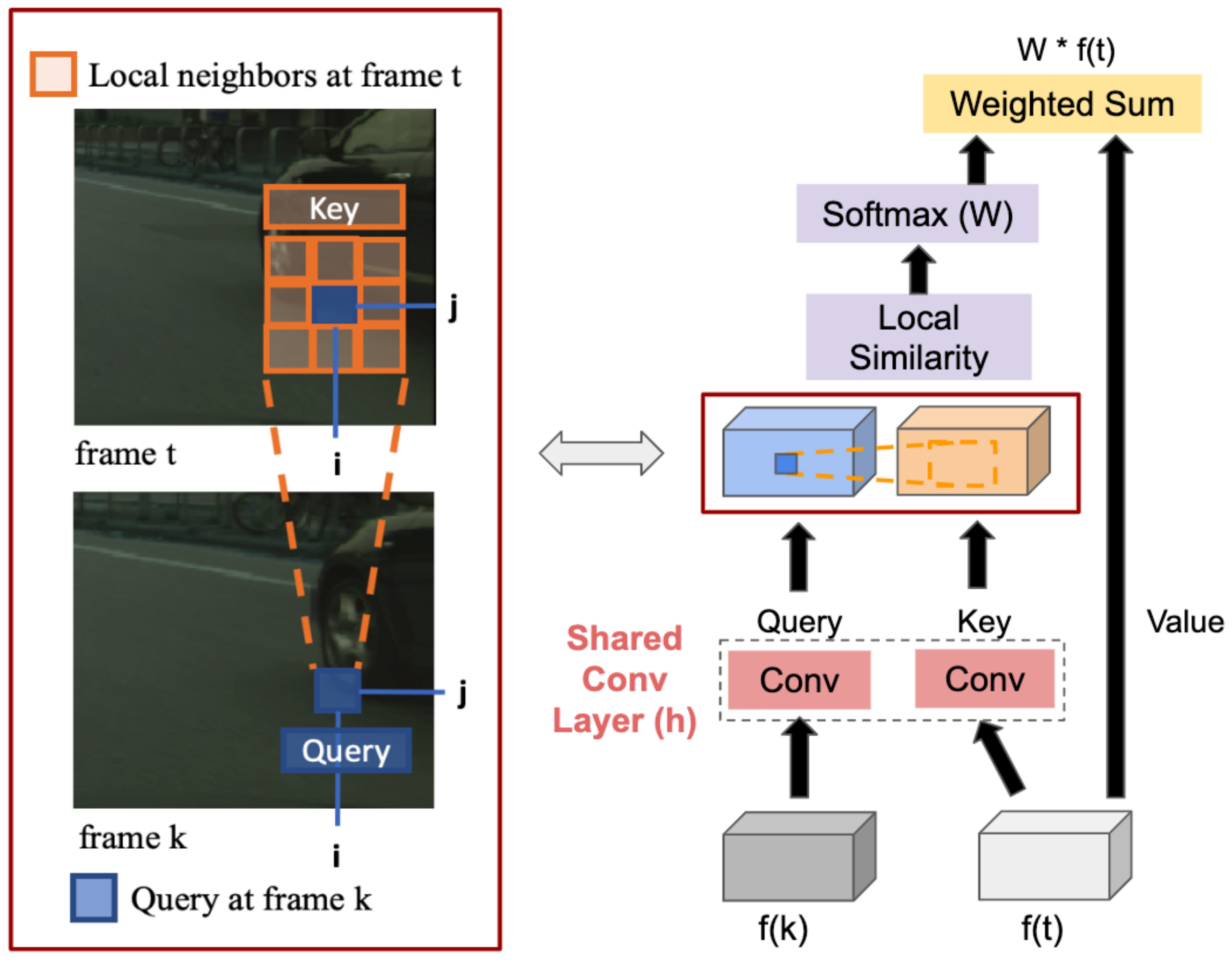} 
    \caption{Inter-frame local attention (ILA) accounts for motion by finding local attention weights in inter-frames. With the shared conv layer, our module generates high attention weighs on the similar features between frames.}
    \label{fig:ILA}
    \vskip -0.2in
\end{figure}

At the decoder step, we perform predictions on each task with a task-specific decoder $\{D_1, D_2, \ldots, D_M\}$, where M is the total number of tasks. Each task-specific decoder consists of squeeze-excitation (SE) blocks on top of shared features from the encoder, inter-frame local attention (ILA) modules to extract and propagate task-specific features across frames and a set of conv layers. In order to fully leverage temporal information, we enable multi-frame feature propagation: a non-keyframe receives features propagated from the last keyframe and the last non-keyframe; a keyframe receives features propagated from the last non-keyframe. This is different from existing feature propagation~\cite{jain2019accel,li2018low} which only propagates features from a keyframe to a non-keyframe. 

\subsection{Inter-Frame Local Attention (ILA)} 
\label{sec:attention}
The key challenge for attention based feature propagation is how to leverage inter-frame temporal cues to propagate features efficiently and effectively. We introduce a light-weight inter-frame local attention (ILA) module for feature propagation. As illustrated in Figure~\ref{fig:ILA}, ILA computes local attention weights $W$ from the feature maps of two different frames (either two neighboring frames or a non-keyframe and a keyframe) to exploit local motion changes. 

\begin{figure}[t]
\vskip -0.1in
		\centering
		\includegraphics[width=0.95\linewidth]{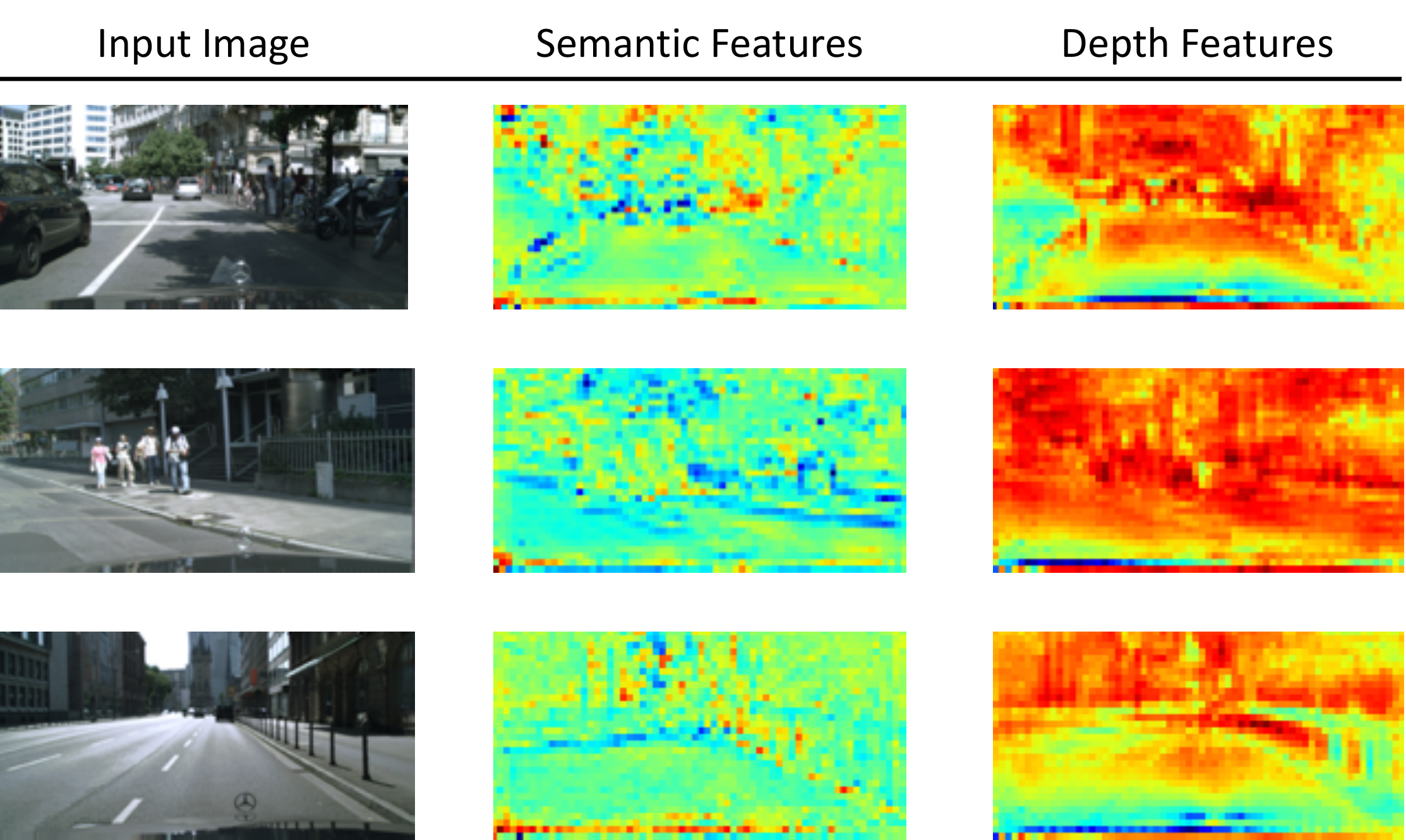} 
		\caption{Visualization of task-specific features from our task-specific attention module.
		}
		\label{fig:task_feature}
\end{figure}
Given a pair of frames $I_t$ and $I_k$, ILA operates on feature maps $f_t$ and $f_{k}$ and propagates features from $f_t$ to $f_k$. In our design (see Figure~\ref{fig:architecture}), the feature maps are the output of task-specific squeeze-and-excitation (SE) blocks~\cite{hu2018squeeze}. For each pixel on the feature map $f_k$, we propagate the features from $f_t$ based on a weighted combination of pixels in a local neighborhood.  
\begin{equation}
\label{eq:attention}
     f_{t\rightarrow k}(i, j) = \sum_{\substack{x=-L/2}}^{L/2}\sum_{\substack{y=-L/2}}^{L/2} W_{i,j}(x,y) f_t(i+x,j+y),
\end{equation}
where $(i,j)$ denotes the pixel location in the image, $L$ is the window size and $W$ is the attention weight obtained by measuring the similarity between the two feature maps $f_t$ and $f_k$. The attention weight matrix $W$ is defined in the following:   
\begin{equation}
\begin{split}
W_{i, j}(x, y) &= \text{softmax}( h(f_{k}(i,j)) \cdot h(f_{t}(i+x,j+y))),
\end{split}
\label{eq:attention2}
\end{equation}
where $W_{i, j}(x, y)$ is the attention weight which measures the similarity between features at position $(i,j)$ and $(i+x, i+y)$ of the two feature maps, respectively. $h$ is a $3\times3$ convolution layer shared between the two feature maps to capture the semantic information in a local window around pixel $(i,j)$. We use inner product to capture the similarities. Then a softmax layer is applied to ensure the sum of weights equals to $1$. Note that ILA is performed only on local neighborhoods, resulting in reduced computational cost as compared to existing global attention modules~\cite{fu2019dual}. 

\subsubsection{Task-specific Attention}~\label{sec:task-specific}
A common challenge in multi-task learning asks how to balance the shared and task-specific features. A heavily shared representation can reduce computational costs and can help prevent over-fitting, but it can also hurt accuracy due to limited model capacity to handle multiple tasks~\cite{maninis2019attentive}. To solve this issue, methods that add extra task specific layers to the multi-task network~\cite{liu2019end,misra2016cross,ruder122019latent} gained popularity during the recent years and achieved higher task accuracy. The drawback is that the complex task-specific layers also significantly increase the computational burden.

Our ILA module is task-specific in order to learn discriminative task-specific features. In contrast to prior work, ILA learns to select and propagate features from previous frames rather than attending the features in the current frame. Leveraging temporal information drastically reduces the required complexity of task-specific layers as the model capacity and discriminative power are shared across multiple frames. Unlike the previous heavy attention modules in  \cite{fu2019dual,liu2019end}, the other advantage is that ILA only attends to features from the task-specific SE blocks within a local window from previous frames. This assumption on temporal consistency reduces the computational cost of ILA. Compared to state-of-the-art attention-based multi-task network~\cite{liu2019end}, MILA achieves better accuracy with 54\% reduction of FLOPs.  

Visualization of the learned task-specific features are shown in Figure~\ref{fig:task_feature}. We can see clear differences in feature patterns for different tasks. Semantic segmentation features highlight object patches, lines and boundaries, while the depth features highlight foreground and background. This confirms the effectiveness of ILA as a feature selector to focus on parts that are discriminative for each task. 

\begin{figure}[t]
		\centering
		\includegraphics[width=1.0\linewidth]{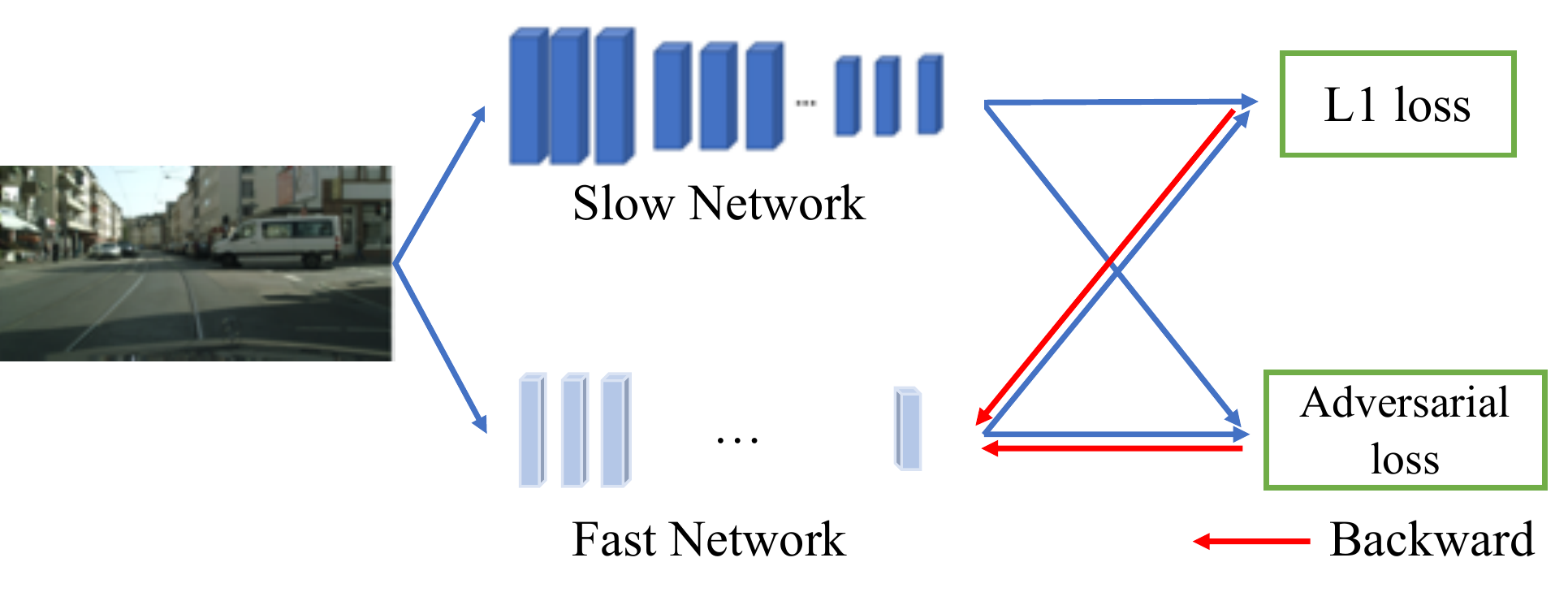} 
		\vskip -0.1in
		\caption{Adversarial learning. In order to let our attention module (ILA) capture temporal consistency, we adopt an adversarial learning strategy for training, where we use a combination of L1 and adversarial loss for the $Fast$ network to mimic the features learned by the $Slow$ network.
		}
		\vskip -0.1in
		\label{fig:mimick}
\end{figure}

\subsection{Boosting ILA for \textit{SlowFast}}
\label{sec:mimicking}
ILA assumes similar features propagate across frames. The high-level idea is similar to optical flow which assumes color constancy between pixels in consecutive images in order to capture motion. However, different backbones (e.g. ResNet-101 and ResNet-18) from the $Slow$ and $Fast$ branches cannot guarantee learning similar features for similar image patches. Thus, naive attention modules could not improve accuracy in our experiments (see Table~\ref{tab:ablation_loss}).

We adopt adversarial learning to train the network so that the $Fast$ network learns similar features to the more accurate $Slow$ network. Figure~\ref{fig:mimick} illustrates our approach. During training, we use a discriminator $D$~\cite{ganin2016domain,Goodfellow2014GenerativeAN} to classify whether the features are output of the $Slow$ network or the $Fast$ network, and the $Fast$ network is trained to confuse the discriminator by ``mimicking'' the output features of the $Slow$ network. In practice, we observed combining $L1$ loss with the adversarial loss lead to improved accuracy. Our loss function $\mathcal{L}$ is defined in the following:
\begin{equation}
\begin{split}
\label{eq:mimic}
    \mathcal{L} & =  \min (\alpha \mathcal{L}_{L1} - \beta \min_{D}\mathcal{L}_{adversarial})\\
    \mathcal{L}_{L1} &= |Slow(I_t) - Fast(I_t)| \\
    \mathcal{L}_{adversarial} &= \log D(Slow(I_t)) + \log (1-D(Fast(I_t))),\\
\end{split}
\end{equation}
where $Slow(I_t)$ and $Fast(I_t)$ are the features of the $Slow$ and $Fast$ backbones on image $I_t$. The loss function $L$ enforces the $Fast$ network to mimic the features learned from the $Slow$ network. $D$ is only used during training and does not increase computation in inference time.




\section{Experiments}
We validate our approach (MILA) in the following two aspects for both accuracy and computation cost. (1) Comparison with the state-of-the-art multi-task learning approaches on videos, the ablation study for our proposed task-specific ILA, and our training losses. (2) The efficacy of our attention based feature propagation approach (ILA) compared with other feature propagation methods.



\begin{table*}[t]
\small
\centering
\resizebox{0.92\textwidth}{!}{\begin{tabular}{|l|c|c|c|c|c|c|c|c|c|c|}
\hline
\multirow{2}{*}{Model}                                       & \multicolumn{2}{c|}{\multirow{2}{*}{Segmentation}} & \multicolumn{2}{c|}{\multirow{2}{*}{Depth}}      & \multicolumn{5}{c|}{Normal Estimation}                & \multirow{2}{*}{GFLOPs}                                 \\ \cline{6-10}
& \multicolumn{2}{c|}{}& \multicolumn{2}{c|}{}  & \multicolumn{2}{c|}{Angle Dist. $\downarrow$} & \multicolumn{3}{c|}{Angle$^{\circ}$ Within $\uparrow$} & \\ \cline{2-10}
& mIOU $\uparrow$            & Acc. $\uparrow$      & Abs. $\downarrow$ & Rel. $\downarrow$ & Mean                 & Median                 & 11.25$^{\circ}$        & 22.5$^{\circ}$       & 30.0$^{\circ}$ &        \\ \hline \hline
D101-SingleTask & 37.2 & 75.0 & 39.3& 16.6 & 22.8 & 16.6 & 35.7 & 62.8 & 73.9 & 236 \\
D101-MultiTask  & 37.1 & \textbf{75.3} & 39.0& 16.3 & 23.7 & 17.6 & 34.0 & 60.2 & 71.7 & 79 \\
MTAN-Seg.~\cite{liu2019end} & 17.7 & 55.3 & 59.0 & 25.8 & 31.4 & 25.4 & 23.2 & 45.7 & 57.6  & 178 \\
Cross-Stitch-Seg.~\cite{liu2019end} & 14.7 & 50.2 & 64.8 & 28.7 & 33.6 & 28.6 & 20.1 & 40.5 & 52.0 & 213\\
MTAN* & 37.1 & 74.3 & 40.0 & 16.9 & 23.9 & 18.1 & 33.5 & 59.5 & 70.4  & 151 \\
Cross-Stitch* & 37.5 & 74.5 & 39.5 & 16.2 & \textbf{22.7} & \textbf{16.5} & \textbf{36.8} & \textbf{63.0} &
 \textbf{73.8} & 236 \\
MultiNet++*~\cite{chennupati2019multinet++} & 32.8& 73.1& 41.1& 17.3 & 24.4 & 18.1 & 33.4 & 58.9 & 70.2 & \textbf{40}   \\
\hline
MILA w/o MF\&TS & 36.6 & 74.8 & 39.2 & 16.5 & 23.5 & 17.4 & 34.2 & 60.4 & 71.8 & 41  \\
MILS w/o TS & 37.0 & 75.0 & 38.9 & 16.4 & 23.3 & 17.2 & 34.9 & 60.9 & 72.2  & 46 \\ 
MILA (Ours) & \textbf{38.1} & 75.1 & \textbf{38.6} & \textbf{16.1} & 23.2 & 17.0 &  35.4 & 61.8 & 72.5 & 70 \\ 
\hline
\end{tabular}}
\vskip -0.1in
\caption{Comparisons for video based Multi-task learning on NYUd v2 dataset. 
* means training with the same Deeplab-ResNet101 backbone as ours. Cross-stitch* shows better results in the normal estimation task mostly because it contains task-specific backbones. Our method obtains high-quality predictions while reducing the massive computational burden.
}
\label{tab:nyu}
\end{table*}

\subsection{Implementation Details}
We implement MILA using PyTorch. We train MILA using ADAM optimizer with $\beta_1 = 0.9$ and $\beta_2=0.99$. The learning rate is $1e^{-4}$ and batch size is 4. The training loss converges after 50 epochs. For the adversarial loss in Eq.~\ref{eq:mimic}, we set $\alpha = \beta=1$. ILA computes on a window size of $L=5$ in Eq.~\ref{eq:attention2}.
We use DeepLab-ResNet101~\cite{chen2017deeplab,He_2016_CVPR,long2015fully} as a $Slow$ network and DeepLab-ResNet18 as a $Fast$ Network. The backbones are pre-trained on ImageNet~\cite{deng2009imagenet} and finetuned for multi-task learning. For brevity, Deeplab-ResNet101 is denoted as D101. D101-18 refers to the \textit{SlowFast} network with DeepLab-ResNet101 and DeepLab-ResNet18. Each task-specific decoder consists of three convolution layers with kernel size of 3x3, 1x1, and 1x1 respectively, and feature size of 512 and 256 in between.
\smallskip

\noindent\textbf{Keyframe Interval.} We train our network with a fixed keyframe interval of $K=5$ following~\cite{jain2019accel}~(every $5$-th frame is a keyframe). For evaluation, since frames in a video are sparsely annotated (\eg~20-th frame in a video clip) in existing datasets, we measure performances of an annotated frame by running our method for all possible keyframe interval offsets $[0, K-1]$ and report the averaged accuracy and GFLOPs. For the evaluation of ILA on semantic video segmentation, we use the same keyframe intervals as the compared methods~($5$ and $10$). 

\subsection{Setup}\label{exp:dataset}

\noindent\textbf{Datasets.} We evaluate our MILA on two widely used public video datasets: Cityscapes~\cite{cordts2016cityscapes} and NYUd v2~\cite{silberman2012indoor}. We follow the evaluation protocols as in Liu~\etal~\cite{liu2019end}: on Cityscapes, we perform \textit{$2$ task predictions} including 7-class segmentation and depth estimation, where images are resized to $256\times512$ to boost up training process; on the NYUd v2 dataset, we perform \textit{$3$ task predictions} including 13-class segmentation, depth estimation, and normal estimation, with input images resized to $288\times384$. For evaluation of ILA on the single task of semantic video segmentation, we perform 19-class segmentation the same as the state-of-the-art by Jain~\etal~\cite{jain2019accel}.
\smallskip

\noindent\textbf{Metrics.} 
For semantic segmentation, we use mean intersection-over union (mIoU) metric and pixel accuracy (PA). For depth estimation, we evaluate on absolute and relative depth errors from the ground truth. For normal estimation, we measure the mean and median angle distances between the predicted angles and ground-truth angles. We also measure the percentage of pixels that are within the angles of $11^{\circ}, 22.5^{\circ}, 30^{\circ}$ to the ground-truth. We compare on computation cost based on GFLOPs following~\cite{feichtenhofer2019slowfast,tang2018experimental,zoph2018learning} and use the “thop” library~\footnote{https://github.com/Lyken17/pytorch-OpCounter} for counting GLFOPs.
\smallskip

\noindent\textbf{Baselines.} We compare with state-of-the-art multi-task learning methods: \textbf{MTAN}~\cite{liu2019end}, \textbf{Cross-Stitch} network~\cite{misra2016cross} and \textbf{MultiNet++}~\cite{chennupati2019multinet++}.  MTAN and Cross-Stitch are single frame based, while MultiNet++ uses multi-frame inputs. Although MTAN and Cross-Stitch used different backbones in their respective papers, for a fair comparison, we report the performance of the two using the same DeepLab-ResNet101 backbone as our $Slow$ network~(which shows better performance than the backbones from their papers). We use the four outputs of each group of layers containing the residual blocks in the backbone as input for the attention modules for the two methods (see Figure~\ref{fig:overview}). 
For our method, and MultiNet++, we use the \textit{SlowFast} network with ResNet101 and ResNet18.  We also compare with two other baselines: \textbf{D101-SingleTask}, which uses a separate ResNet101 backbone for training each task without any shared features; \textbf{D101-MultiTask}, which uses the shared ResNet101 backbone with task-specific decoders. 

\begin{figure*}[t]
		\centering
		\includegraphics[width=0.9\linewidth]{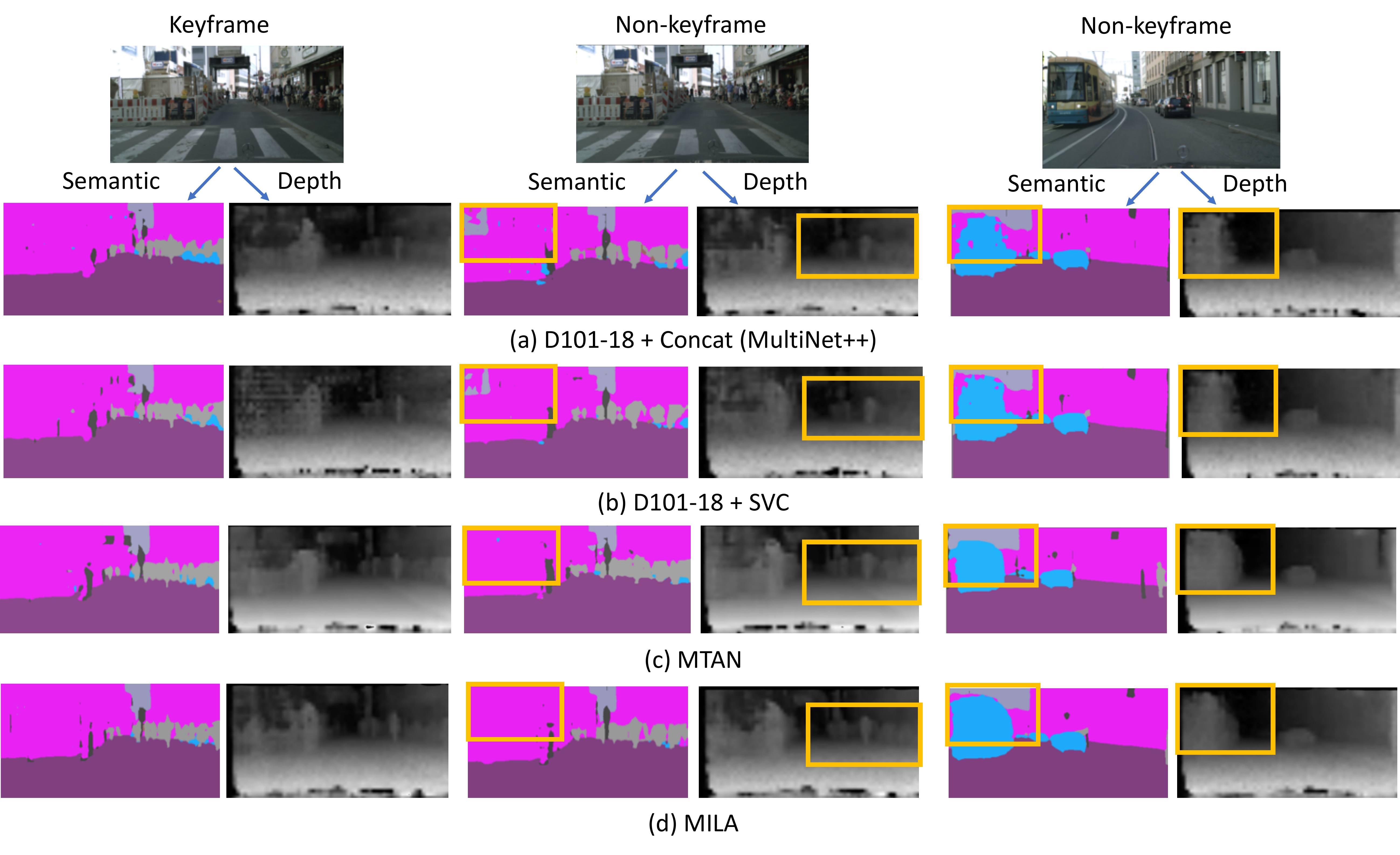} 
		\vskip -0.1in
		\caption{Qualitative results on Cityscapes. We choose a keyframe and non-keyframe with offset 4.  For the keyframe, the three methods produce very similar qualitative results. For non-keyframes, the baseline methods (a, b) performs worse (see orange boxes) but our method (d) still obtains robust performance even compared with the $Slow$-only network (c). }
		\label{fig:qualitative}
\end{figure*}

\begin{table}[t]
\centering
\small
\setlength{\tabcolsep}{2.5pt}
\resizebox{0.45\textwidth}{!}{\begin{tabular}{|l|c|c|c|c|c|}
\hline
\multirow{2}{*}{Model}                                                  & \multicolumn{2}{c|}{Segmentation} & \multicolumn{2}{c|}{Depth} & \multirow{2}{*}{GFLOPs} \\ \cline{2-5}
                                                                        & mIOU $\uparrow$       & Acc. $\uparrow$        & Abs.  $\downarrow$     & Rel. $\downarrow$   & \\ \hline \hline

D101-SingleTask & 63.9 & 94.4 & \textbf{1.02} & 25.3 & 187    \\
D101-MultiTask & 63.8 & 94.4 & 1.06 & 31.9 & 93 \\
                        
MTAN-SegNet~\cite{liu2019end}  & 53.0 & 91.1 & 1.44 & 33.6 & 168 \\
MTAN * & 64.2 & 94.5 & 1.06 & 26.3 & 161 \\
Cross-Stitch* & \textbf{64.5} & 94.5 & 1.04 & 33.0 & 187 \\
MultiNet++~\cite{chennupati2019multinet++} & 61.6& 93.9& 1.08& 28.5 & \textbf{47} \\
\hline
MILA w/o MF\&TS & 63.8 & 94.4 & 1.05 & 32.9 & 48\\

MILA w/o TS& 64.1 & 94.5 & 1.03 & 31.5 & 55\\
MILA (Ours) & 64.3 & \textbf{94.6} & \textbf{1.02} & \textbf{25.2} & 70\\
\hline
\end{tabular}}
\vskip -0.1in
\caption{Comparison for video based multi-task learning on the Cityscapes dataset. * means training with the same Deeplab-ResNet101 backbone. Ours and MultiNet++ use the D101-18 backbone.}
\label{tab:cityscapes}
\vskip -0.2in
\end{table}


\subsection{Video Based Multi-Task Learning}
\noindent\textbf{Quantitative results.} We report the performance of video based multi-task learning on the NYUd v2 dataset in Table~\ref{tab:nyu} and the Cityscapes dataset in Table~\ref{tab:cityscapes} respectively. In Table~\ref{tab:nyu}, on the NYUd v2 dataset, we outperform other approaches for depth estimation and one metric of mIOU for semantic segmentation, with ranked 2nd segmentation accuracy. MILA shows slightly worse performance for normal estimation than Cross-Stitch. This is because Cross-Stitch has task-specific backbones, while MILA use a single shared one, which is significantly computationally efficient, saving $70\%$ of computational costs (236 vs. 70 GFLOPs) compared to Cross-Stitch~\cite{misra2016cross}. MILA also saves 46\% (70/151 GFLOPs) of computations compared to MTAN~\cite{liu2019end} as shown in Table~\ref{tab:nyu} by replacing task-specific heavy layers with light-weight task-specific ILAs. We rank the 2nd for GFLOPs right after MultiNet++ but with much better accuracy. In Table~\ref{tab:cityscapes}, MILA outperforms all other methods for depth estimation and achieves the the best accuracy for semantic segmentation, while ranking the 2nd for mIOU on Cityscapes. 
\smallskip

\noindent\textbf{Qualitative results.} We show in Figure~\ref{fig:qualitative} sample qualitative results from the Cityscapes dataset. Baselines with a \textit{SlowFast} model (a,b) decreases accuracy and outputs noisy predictions in non-keyframes compared to a keyframe. But our method (d) produces robust predictions on non-keyframes even compared with MTAN (c) which uses $Slow$-only network. Please check more results and analyses in our supplementary material. 
\smallskip
\begin{table}[t]
\centering
\small
\begin{tabular}[t]{|l|ccc|cc|}
\hline
Backbone & L1 & Adv. & ILA & mIOU ($\uparrow$) & Depth Err. ($\downarrow$) \\
\hline
\hline
D101-18 &  & &  & 61.6 & 1.08 \\
D101-18 & \checkmark &  &  & 61.8 & 1.07 \\
D101-18 & &  & \checkmark & 61.9 & 1.08  \\
D101-18 &  \checkmark &   & \checkmark & 63.3 & \textbf{1.05}\\
D101-18 &  \checkmark & \checkmark  & \checkmark & \textbf{63.8} & \textbf{1.05}\\
\hline
\end{tabular}
\vspace{-0.1in}
\caption{Ablation study of ILA on Cityscapes. ILA with L1 loss and adversarial loss (denoted by Adv.) leads to clear improvement. ILA without the proposed losses obtains similar performances as just simple concatenation~\cite{chennupati2019multinet++}. These losses do not increase GFLOPs in inference time.}
\label{tab:ablation_loss}
\vspace{-0.2in}
\end{table}

\noindent\textbf{Ablation study.} The last three rows in Tables~\ref{tab:nyu} and ~\ref{tab:cityscapes} show the ablation study of MILA method. \textbf{``MILA w/o TS''} means our approach without task-specific attention design. \textbf{``MILA w/o MF\&TS''} means without both task-specific design and feature propagation for neighboring frames. \textbf{``MILA w/o MF\&TS''} achieves similar performance as D101-MultiTask while saving 48\% computations. MILA outperforms D101-MultiTask and reduces 25\% of computations. MILA approach shows the best performed accuracy with a small increase in GFLOPs. 

We show the ablation study on our inter-frame local attention (ILA). Table~\ref{tab:ablation_loss} reports the impact of the adversarial and L1 losses (Eq.~\ref{eq:mimic}). In the third line of  Table~\ref{tab:ablation_loss}, it shows that the local attention module alone does not greatly improve the performances. However, when ILA is combined with the proposed losses, it significantly increases the accuracy. In addition, we show the ablation study for the local window size in Table~\ref{tab:ablation_ILA}. ILA is not sensitive to small changes in the window size, but performance drops significantly when the window size is global. In addition, a small window size is faster as it attends to less number of neighbors than the large window size.

\begin{table}[t]

\centering
\begin{tabular}[t]{|l|c|c|c|}
\hline
Backbone & Window Size & mIOU ($\uparrow$) & Depth Err. ($\downarrow$) \\
\hline
\hline
D101-18 & 3x3         &    64.1  & \textbf{1.02}   \\
D101-18 & 5x5         &  \textbf{64.3}   &    \textbf{1.02}  \\
D101-18 & 7x7         &  64.2    &  \textbf{1.02}  \\
D101-18 & 15x15         &  64.0    & 1.04  \\
D101-18 & Global (64)      &  62.4    & 1.10   \\
\hline
\end{tabular}
\vspace{-0.1in}
\caption{Ablation study on the kernel size in ILA on Cityscapes. ILA performs better than global attention for feature propagation.}
\label{tab:ablation_ILA}
\end{table}

\begin{table}[]
\centering
\begin{tabular}{|l|c|l|c|}
\hline
Backbone & $K$ & Feature Prop. & mIOU (\%) \\
\hline
\hline
D101-18 & 5 & Optical flow & 72.1 \\
D101-18 & 5  & ILA (Ours) & \textbf{73.2} \\
\hline
D101-18 & 10  & Optical flow & 69.8 \\
D101-18 & 10  & ILA (Ours) & \textbf{72.1} \\
\hline
D101-34 & 5 & Optical flow & 72.4 \\
D101-34 & 5 & ILA (Ours) & \textbf{74.3} \\
\hline
D101-34 & 10 & Optical flow & 70.1 \\
D101-34 & 10 & ILA (Ours) & \textbf{73.8} \\

\hline
\end{tabular}
\vspace{-0.1in}
\caption{Comparison with optical-flow based feature propagation~\cite{jain2019accel} for the semantic segmentation task on Cityscapes. A keyframe interval is denoted by $K$.}
\label{tab:semantic_segmentation}
\end{table}

\subsection{Detailed Analysis on Feature Propagation}\label{sec:FP_comparison}
We compare ILA with the two feature propagation methods:(1) optical flow based warping with FlowNet-S~\cite{dosovitskiy2015flownet}, which shows state-of-the-art performance on the single task of semantic video segmentation for the Accel method~\cite{jain2019accel} and (2) spatially variant convolution layers (SVC)~\cite{li2018low} which is proposed to propagate features from a keyframe to non-keyframes. 
We use the same backbone for all methods for a fair comparison.
\smallskip

\noindent\textbf{Performance comparison.} To compare with optical flow based warping~\cite{jain2019accel}, we follow their semantic segmentation evaluation protocol in Table~\ref{tab:semantic_segmentation}. Feature propagated with ILA obtains higher accuracy than optical flow-based warping~\cite{jain2019accel}. We observe that the accuracy improvements of ILA are more evident in the higher keyframe interval. 

In Table~\ref{tab:FP_compare_multi}, we provide a comparison with SVC~\cite{li2018low} on Cityscapes and NYUd v2 on multi-task learning. ILA outperforms all other methods in the two datasets while requiring less computational burden. We observe that the quality of optical flow estimation is poor in this evaluation protocol (\ie~low-resolution images), which results in significantly worse performance. 
\smallskip

\begin{table}[t]
\centering
\small
\begin{tabular}{|rl|c|c|c|c|}
\hline
&\multirow{2}{*}{Feature Prop.}                                                  & \multicolumn{2}{c|}{Segmentation} & \multicolumn{2}{c|}{Depth}  \\ \cline{3-6} & & mIOU $\uparrow$       & Acc. $\uparrow$        & Abs.  $\downarrow$     & Rel. $\downarrow$    \\ \hline \hline
{\bf (a)}& Cityescapes & & & & \\                                                           
&SVC~\cite{li2018low}& 62.3& 94.0& 1.06& 33.3\\
&ILA (Ours)& \textbf{63.8}& \textbf{94.4}& \textbf{1.05}& \textbf{32.9}\\
\hline
{\bf (b)}& NYUv2  & & & & \\                                  
&SVC~\cite{li2018low}& 35.7&74.7&40.3& 17.0  \\
&ILA (Ours) & \textbf{36.6}& \textbf{74.8}& \textbf{39.2}& \textbf{16.5}  \\
\hline
\end{tabular}
\vspace{-0.1in}
\caption{Comparison with feature propagation methods with D101-18 backbone on Cityscapes and NYUv2}
\label{tab:FP_compare_multi}
\end{table}

\begin{table}[t]
\centering
\small
\begin{tabular}{|l|c|c|c|}
\hline
Feature Propagation & GFLOPs & \# Conv. & \# Param \\
\hline
\hline
(a) Input size: $258\times512$ & & & \\
Optical flow ~\cite{dosovitskiy2015flownet,jain2019accel}  & 7.5    & 23                          & 38M     \\
SVC ~\cite{li2018low} & 5.4    & 3                           & 3M      \\
ILA (Ours)              & \textbf{0.2}    & \textbf{1}                           & \textbf{0.2M}   \\
\hline
(b) Input size: $1024\times2048$ & & & \\
Optical flow ~\cite{dosovitskiy2015flownet,jain2019accel}  & 71.2    & 23                          & 38M     \\
SVC ~\cite{li2018low} & 108    & 3                           & 3M      \\
ILA (Ours)    & \textbf{5.4}    & \textbf{1}                           & \textbf{0.2M}   \\
\hline
\end{tabular}
\vspace{-0.1in}
\caption{Comparison on feature propagation modules. SVC represents the method of~\cite{li2018low}. Our method is light-weight and computationally efficient.}
\label{tab:FP_comparison}
\end{table}

\noindent\textbf{Space and computation cost.} In Table~\ref{tab:FP_comparison}, we show the comparison of GFLOPs, the number of convolutional layers and the number of parameters for the optical flow warping, SVC, and our ILA. We report numbers given different input sizes. ILA consists of only one convolutional layer, making it much more memory efficient than the other two methods. For GFLOPs, other methods require more computations than ILA and the gain is more evident when the input size is larger. ILA takes only 4\% (0.2/5.4 GFLOPs) of computations in the SVC feature propagation~\cite{li2018low}. 
Additional analyses can be found in our supplementary material.

\section{Conclusion}
We present an efficient and effective multi-task learning framework on video streams. We propose a novel task-specific inter-frame local attention (ILA) module, which accounts for motion and propagate discriminative task-specific features over time in a spatial-variant manner. Our attention module is much faster, more accurate, and modular compared to prior feature propagation methods. Our inter-frame local attention module can be used to extract task-specific features with minimal computation compared to existing heavy task-specific layers. Our experiments show that our method significantly reduces the computational cost without compromising accuracy compared to the state-of-the-art multi-task learning models.

{\small
\bibliographystyle{ieee_fullname}
\bibliography{egbib}
}

\appendix

\section*{Appendix}

\section{Experiment Details for Fair Comparison}
As stated in Sec.4.2 in the main paper, for fair comparison to other methods (MTAN~\cite{liu2019end}, Cross-Stitch~\cite{misra2016cross} and MultiNet++~\cite{chennupati2019multinet++}), we unify the backbone for all method as Deeplab-ResNet101~\cite{chen2017deeplab,He_2016_CVPR}. Note that after using Deeplab-ResNet101, the performance for MTAN, Cross-Stitch and MultiNet++ gets improved compared to their originally reported numbers as shown in Table~\ref{tab:cityscapes_sup}. Deeplab-ResNet101 is initialized with pre-trained weights on ImageNet but decoders use randomly initialized weights. Then, the model is finetuned for multi-task learning.

For implementation details, since all multi-task learning models except Cross-Stitch use shared encoder, we pre-train a shared multi-task encoder based on Deeplab-ResNet101 for each method (named as``D101-MultiTask''). For the Cross-Stitch model we pre-train a task specific encoder~(named as ``D101-SingleTask''). Using networks pre-trained on multi-task learning shows better performances than just directly finetuning pre-trained weights on ImageNet~\cite{gao2019nddr}. We first train D101-SingleTask and D101-MultiTask with Adam optimizer with learning~\cite{kingma2014adam} rate $1e^{-4}$ with a decay rate $1e^{-1}$. For our SlowFast based network, we use the D101-MultiTask for our slow network. We also train a D18-MultiTask for the $Fast$ network with the same optimizer and training setting. For fair comparison with MultiNet++, we use the same pre-trained slow-fast network for MultiNet++, but concatenate neighboring frames as input to learn from videos as proposed in their paper.

Note that we also report in the main paper the performance comparison with our pre-trained ``D101-SingleTask'' and ``D101-MultiTask''.

\begin{table}[t]
\centering
\small
\setlength{\tabcolsep}{3.0pt}
\begin{tabular}{|l|c|c|c|c|}
\hline
\multirow{2}{*}{Model}                                                  & \multicolumn{2}{c|}{Segmentation} & \multicolumn{2}{c|}{Depth}  \\ 
\cline{2-5}
& mIOU $\uparrow$ & Acc. $\uparrow$ & Abs.  $\downarrow$ & Rel. $\downarrow$  \\ \hline \hline

MTAN-SegNet~\cite{liu2019end}  & 53.0 & 91.1 & 1.44 & 33.6  \\
MTAN-ResNet101 & \textbf{64.2} & \textbf{94.5} & \textbf{1.06} & \textbf{26.3}  \\
Cross-Stitch-SegNet~\cite{liu2019end} & 50.1 & 90.3 & 1.54 & 34.5\\
Cross-Stitch-ResNet101 & \textbf{64.5} & \textbf{94.5} & \textbf{1.04} & \textbf{33.0} \\
\hline
\end{tabular}
\caption{Comparison of different backbones. Using the ResNet101 backbone improves the accuracy by a large margin.}
\label{tab:cityscapes_sup}
\end{table}

\subsection{Weighting Strategies.} In the main paper, we use the equal weighting for each task. In Table~\ref{tab:weighting}, we show performance with different weighting strategies for the multiple tasks: Uncertainty Weighting~\cite{kendall2018multi} and Dynamic Weight Average~\cite{liu2019end}. These weighting strategies are proposed to find a balance between different tasks, since a model can be biased to a certain task. A desired multi-task learning model should not depend on these weighting schemes, so that the model itself can find a proper balance between tasks. We observe that MILA achieves the similar performance on different weighting strategies and thus is not sensitive to the weighting schemes.

\begin{table}
\centering
\small
\begin{tabular}[t]{|l|c|c|c|}
\hline
Backbone & Weighting & mIOU ($\uparrow$) & Depth Err. ($\downarrow$) \\
\hline
\hline
D101-18 & Equal         &   \textbf{64.3}   &    1.02        \\
D101-18 & Uncertainty~\cite{kendall2018multi}         &   \textbf{64.3} & \textbf{1.01}  \\
D101-18 & DWA~\cite{liu2019end}         &  64.2    &  1.02 \\
\hline
\end{tabular}
\caption{Different weighting strategies for MILA. MILA is not sensitive to the weighting strategies.}
\label{tab:weighting}
\end{table}

\section{Ablation for $Slow$ network-only}
We apply our full feature propagation method (inter-frame local attention (ILA) + multi-frame feature propagation + task specific) on $Slow$ network-only model. This means that the keyframe interval is 1. Table~\ref{tab:slow-only} shows the results on Cityscapes. In all cases, using our feature propagation method shows significant improvements on the depth estimation task.
\begin{table}[h]
\centering
\small
\begin{tabular}{|l|c|c|c|c|}
\hline
\multirow{2}{*}{Model} & \multicolumn{2}{c|}{Segmentation} & \multicolumn{2}{c|}{Depth}  \\ \cline{2-5}
& mIOU $\uparrow$ & Acc. $\uparrow$ & Abs.  $\downarrow$ & Rel. $\downarrow$    \\ \hline \hline

D101-Multi & 63.8 & 94.4 & 1.06 & 31.9 \\
D101-Multi + Ours & \textbf{64.4} & \textbf{94.6} & \textbf{1.02} & \textbf{25.8} \\
\hline
D50-Multi & 63.1 & 94.2 & 1.09 & 33.0 \\
D50-Multi + Ours & \textbf{63.3} & \textbf{94.5} & \textbf{1.02} & \textbf{25.1} \\
\hline
D18-Multi & 60.3 & 93.4 & 1.21 & 34.8 \\
D18-Multi + Ours & \textbf{61.5} & \textbf{93.8} & \textbf{1.07} & \textbf{25.8} \\
\hline
\end{tabular}
\caption{Evaluation of a Slow-only network with our task-specific inter-frame local attention (ILA) module on Cityscapes.}
\label{tab:slow-only}
\end{table}

\section{Additional Comparison for Feature Propagation}
In Table~\ref{tab:semantic_segmentation_sup}, we provide more detailed comparisons between our attention based feature propagation and the optical flow based feature warping method by Jain~\etal~\cite{jain2019accel}. We compare the task of video semantic segmentation as in~\cite{jain2019accel}. Our method outperforms~\cite{jain2019accel}, and is more robust to different keyframe interval. Note that we only apply our ILA module for fair comparison with other feature propagation methods (\ie without multi-frame and task-specific attention).

\begin{table}[h]
\centering
\begin{tabular}{|l|c|l|c|}
\hline
Backbone & $K$ & Feature Prop. & mIOU (\%) \\
\hline
\hline

D101-50 & 5 & Optical flow & 74.2 \\
D101-50 & 5 & ILA (Ours) & \textbf{75.1}\\
\hline
D101-50 & 10 & Optical flow & 72.9 \\
D101-50 & 10 & ILA (Ours) & \textbf{74.8} \\
\hline
\end{tabular}

\caption{Supplements to Table~5 in the main paper. Comparison with optical-flow based feature propagation~\cite{jain2019accel} for the semantic segmentation task on Cityscapes. A keyframe interval is denoted by $K$. We show detailed comparison using different backbones and keyframe intervals.}
\label{tab:semantic_segmentation_sup}
\end{table}

\section{Additional Qualitative Results}
\begin{figure*}[t]
		\centering
		\includegraphics[width=1.0\linewidth]{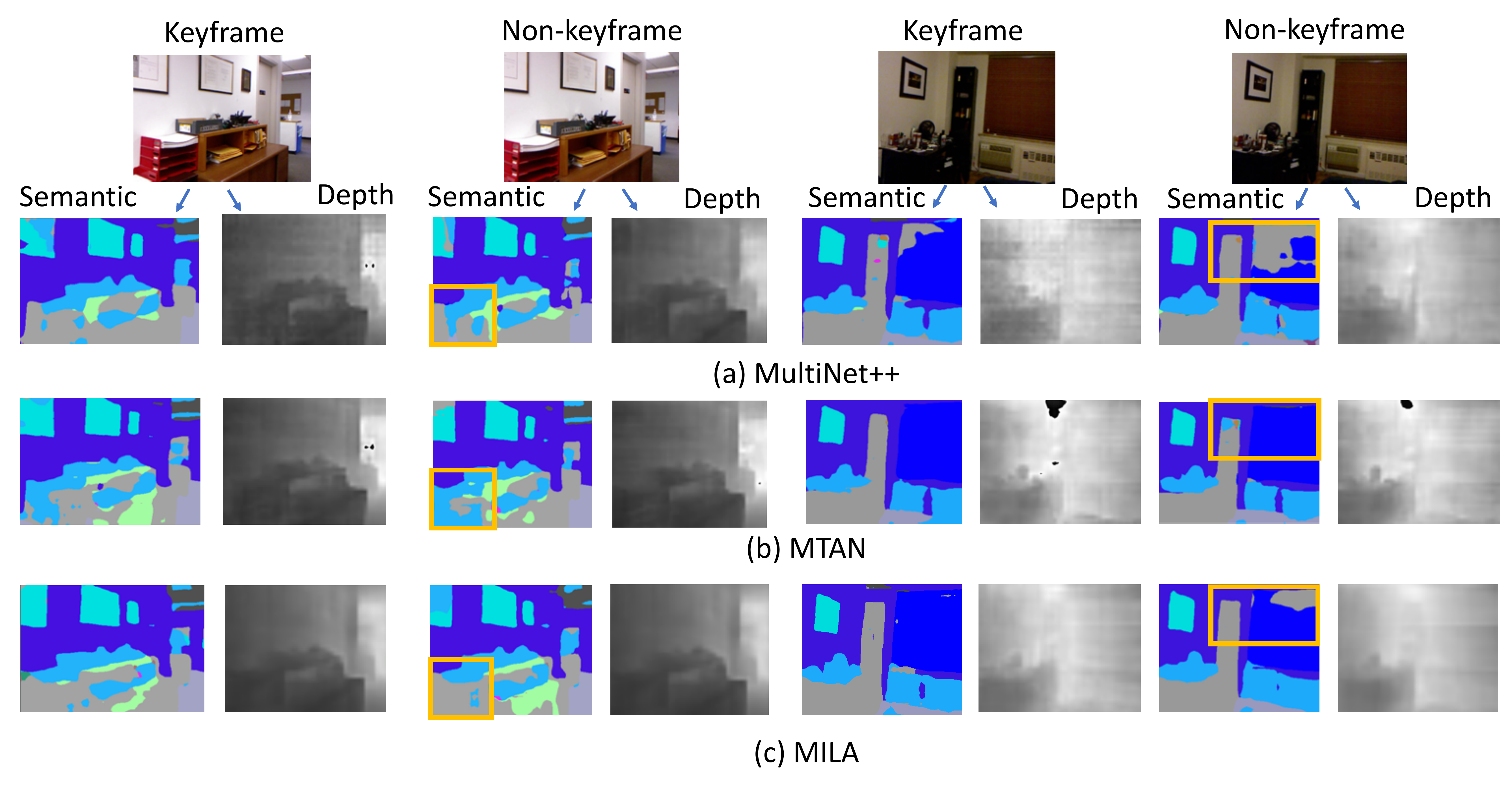} 
		\vskip -0.1in
		\caption{Qualitative results for multi-task learning on videos on the NYUd v2 dataset. We show comparison between (a) MultiNet++, (b) MTAN and (c) ours. MultiNet++ performs worse on the non-keyframe. }
		\label{fig:qualitative_nyu}
\end{figure*}

In the main paper, we only put the qualitative results on Cityscapes due to the limited space. We also provide the qualitative results on NYU v2 in Figure~\ref{fig:qualitative_nyu}. Compared to MultNet++~\cite{chennupati2019multinet++}, our method is robust to non-keyframes and obtains similar performances as the computationally heavy method~\cite{liu2019end}. In addition, we also provide the supplemental video of predictions of the $Slow$ and $Fast$ networks, where the $Fast$ network produces qualitatively similar performances as the $Slow$ network.

\section{Discriminator Details}
The discriminator $D$ takes feature maps from the last residual block of slow and fast network. The discriminator consists of three 3x3 convolutional layers with relu activtaion, a global average pooling layer, and sigmoid activation, which outputs a single value. It should be noted that the discriminator is only used during training, therefore it does not increase GFLOPs in inference time. [R4] We update the encoder and discriminator jointly by using a gradient-reversal layer~\cite{ganin2016domain} for Eq.~3.

\end{document}